%% file: main.tex
\documentclass[a4paper,fleqn]{cas-dc}
\usepackage[utf8]{inputenc}
\usepackage[numbers]{natbib}
\usepackage[T1]{fontenc}
\usepackage{lineno}
\usepackage{hyphenat}
\modulolinenumbers[5]
\usepackage[dvipsnames]{xcolor}
\usepackage{xspace}
\usepackage{csvsimple}
\usepackage{longtable}
\usepackage{lscape}
\usepackage{array}
\usepackage{subfig}
\usepackage{array, etoolbox}
\usepackage[linesnumbered]{algorithm2e}
\usepackage{color}
\definecolor{pblue}{rgb}{0.13,0.13,1}
\definecolor{pgreen}{rgb}{0,0.5,0}
\definecolor{pred}{rgb}{0.9,0,0}
\definecolor{pgrey}{rgb}{0.46,0.45,0.48}
\usepackage[scaled]{beramono}
\usepackage{tikz}
\usepackage{xspace}
\usepackage{listings}
\usepackage[printonlyused]{acronym}

\usepackage{listings}
\definecolor{bgcolor}{rgb}{0.95,0.95,0.92}
\lstdefinestyle{mystyle}{
    backgroundcolor=\color{bgcolor},
    commentstyle=\color{blue},
    keywordstyle=\color{magenta},
    numberstyle=\tiny\color{black},
    stringstyle=\color{black},
    basicstyle=\ttfamily\footnotesize,
    breakatwhitespace=false,
    breaklines=true,        
    captionpos=b,           
    keepspaces=true,        
    numbers=left,           
    numbersep=5pt,          
    showspaces=false,       
    showstringspaces=false,
    showtabs=false,         
    tabsize=2
}

\makeatletter
\newenvironment{btHighlight}[1][]
{\begingroup\tikzset{bt@Highlight@par/.style={#1}}\begin{lrbox}{\@tempboxa}}
{\end{lrbox}\bt@HL@box[bt@Highlight@par]{\@tempboxa}\endgroup}
\newcommand\btHL[1][]{%
  \begin{btHighlight}[#1]\bgroup\aftergroup\bt@HL@endenv%
}
\def\bt@HL@endenv{%
  \end{btHighlight}%
  \egroup
}
\newcommand{\bt@HL@box}[2][]{%
  \tikz[#1]{%
    \pgfpathrectangle{\pgfpoint{1pt}{0pt}}{\pgfpoint{\wd #2}{\ht #2}}%
    \pgfusepath{use as bounding box}%
    \node[anchor=base west, fill=orange!30,outer sep=0pt,inner xsep=1pt, inner ysep=0pt, rounded corners=3pt, minimum height=\ht\strutbox+1pt,#1]{\raisebox{1pt}{\strut}\strut\usebox{#2}};
  }%
}
\lstdefinestyle{Java}{
    language={Java}, basicstyle=\ttfamily\footnotesize, 
    moredelim=**[is][{\btHL[fill=red!17,thin]}]{`}{`},
    moredelim=**[is][{\btHL[fill=green!17,thin]}]{£}{£},
    moredelim=**[is][{\btHL[fill=yellow!17,thin]}]{~}{~},
    moredelim=**[is][{\color{green!17}\btHL[fill=green!17,thin]}]{¤}{¤},
    moredelim=**[is][{\color{red!17}\btHL[fill=red!17,thin]}]{µ}{µ},
}

\usepackage{dsfont}
\usepackage{lineno}
\usepackage[T1]{fontenc}
\usepackage[flushleft]{threeparttable}
\usepackage{multicol}
\usepackage{xspace}
\usepackage{amssymb}
\usepackage{amsthm}
\usepackage{lscape}
\usepackage{multirow}
\usepackage{array}
\usepackage{paralist}
\usepackage{makecell}
\usepackage{graphicx}
\usepackage{tabu}
\usepackage[framemethod=tikz]{mdframed}

\mdfdefinestyle{mpdframe}{
    frametitlebackgroundcolor   =black!15,
    frametitlerule              =true,
    roundcorner                 =5pt,
    middlelinewidth             =0.8pt,
    innermargin                 =0.2cm,
    outermargin                 =0.2cm,
    innerleftmargin             =0.2cm,
    innerrightmargin            =0.2cm,
    innertopmargin              =0.2cm,
    innerbottommargin           =0.2cm
}

\usepackage[colorinlistoftodos,prependcaption]{todonotes}
\newboolean{showcomments}
\setboolean{showcomments}{true}
\ifthenelse{\boolean{showcomments}}
 { \newcommand{\mynote}[2]{
      \fbox{\bfseries\sffamily\scriptsize#1}
        {\small$\blacktriangleright$\textsf{\textcolor{red}{{\em #2}\bf }}$\blacktriangleleft$}}}
        { \newcommand{\mynote}[2]{}}
\usepackage{tabularx}

\usepackage{booktabs}

\definecolor{mymauve}{rgb}{0.58,0,0.82}
\definecolor{mygrey}{rgb}{0.43, 0.5, 0.5}

\usepackage{pbox}

\input{util.tex}

\newcounter{rowcount}
\usepackage{amsthm}
\theoremstyle{definition}

\usepackage[framemethod=tikz]{mdframed}
\mdfdefinestyle{mpdframe}{
    frametitlebackgroundcolor   =black!15,
    frametitlerule              =true,
    roundcorner                 =5pt,
    middlelinewidth             =1pt,
    innermargin                 =0.2cm,
    outermargin                 =0.2cm,
    innerleftmargin             =0.2cm,
    innerrightmargin            =0.2cm,
    innertopmargin              =0.2cm,
    innerbottommargin           =0.2cm
}

\usepackage{url}
\AtBeginDocument{%
}

\usepackage{hyperref}

\begin{document}
\hyphenation{de-compi-ler}

\let\WriteBookmarks\relax
\def\floatpagepagefraction{1}
\def\textpagefraction{.001}
\newcommand\mytitle{Interpretation of Swedish Sign Language using Convolutional Neural Networks and Transfer Learning}
\shorttitle{\mytitle}
\shortauthors{Halvardsson, et~al.}

\title [mode = title]{\mytitle}

\author{Gustaf Halvardsson}[]\ead{gustafha@kth.se}

\author{Johanna Peterson}[]\ead{jpet6@kth.se}

\author{C\'esar Soto-Valero}[orcid=0000-0003-0541-6411]\ead{cesarsv@kth.se}\cormark[1]

\author{Benoit Baudry}[orcid=0000-0002-4015-4640]\ead{baudry@kth.se}

\cortext[cor1]{Corresponding author}

\address[]{KTH Royal Institute of Technology, SE-100 44 Stockholm, Sweden}

\begin{abstract}
The automatic interpretation of sign languages is a challenging task, as it requires the usage of high level vision and high level motion processing systems for providing accurate image perception. In this paper, we use Convolutional Neural Networks (CNNs) and transfer learning in order to make computers able to interpret signs of the Swedish Sign Language (SSL) hand alphabet. Our model consist of the implementation of a pre-trained InceptionV3 network, and the usage of the mini-batch gradient descent optimization algorithm. We rely on transfer learning during the pre-training of the model and its data. The final accuracy of the model, based on 8 study subjects and 9,400 images, is 85\%. Our results indicate that the usage of CNNs is a promising approach to interpret sign languages, and transfer learning can be used to achieve high testing accuracy despite using a small training dataset. Furthermore, we describe the implementation details of our model to interpret signs as a user-friendly web application.
\end{abstract}

\begin{keywords}
Sign language interpretation \sep Machine learning \sep Convolutional neural networks \sep Transfer learning 
\end{keywords}



\maketitle

\input{body.tex}

\bibliographystyle{cas-model2-names}
\bibliography{bibliography/ref.bib}

\end{document}

%% file: body.tex
\section{Introduction}\label{sec:introduction}

In 2018 it was estimated that 466 million people worldwide had a disabling hearing loss~\cite{WHO2018}. When a person in a family turns deaf or is born with impaired hearing, several problems might emerge~\cite{Paul2018}. In particular, deaf people whom often use sign language to communicate, are in many cases dependent on interpreters when, for example, seeking for care. People who need to use sign language are often unable to communicate effectively with people who are not familiar with sign language~\cite{Emmorey2002}. In this context, an application that automatically translates sign language is beneficial since it can improve deaf people's quality of life, especially in terms of increased social inclusion and individual freedom. 

The problem is that just as with spoken languages, all sign languages differ. There is no global sign language shared over the world~\citep{Emmorey2002}. Therefore, a generic translating solution is not enough in order to address this problem for all deaf people in the world. To the best of our knowledge, there are today no  application to help them interpret generic sign language to text. If there would be only one, for example, a tool interpreting only American Sign Language (ASL) would not work on Swedish Sign Language (SSL). Therefore, a generic software application, based on a solution that can easily be adjusted for interpreting many different sign languages, would be a preferable solution. This solution should work automatically, translating sign language to text independently of the physical characteristics of the user. 

An application to interpret sign language could benefit from novel Artificial Intelligence (AI) techniques. AI is the science and engineering of building intelligent machines that can be fed raw data to learn on. The machines can make decisions in situations that they have not encountered before. Machine Learning (ML) is  a large sub-area of AI that specializes in recognizing patterns in data to continuously learn from feedback~\citep{Ray2019}. One commonly used ML architecture is neural networks. A neural network is built up of several layers of artificial neurons~\citep{Kaur2012} that try to mimic the function and behaviour of biological neurons~\citep{Rosen1962}. Each layer of neurons specializes in detecting different features; one layer could for example learn to detect edges when analyzing images. Since the differences between signs in sign languages largely consist of different patterns like hand movements and shiftings, using artificial neural networks that learn from datasets of sign images is useful to develop a model able to detect and interpret signs~\citep{Shi2019}.

Training a neural network to perform this type of image recognition adequately requires a large amount of data~\cite{Atkin1997, Kim3017}. Regarding sign language, the only datasets available are based on the Sign Language MNIST image dataset \footnote{\url{https://www.kaggle.com/datamunge/sign-language-mnist}}, which is based on the American Sign Language\footnote{\url{https://www.kaggle.com/grassknoted/asl-alphabet}}. There are none based on the Swedish Sign Language (SSL) and thus not enough SSL data is available to train a model on. While a few databases of SSL exist (i.e., the SSL dictionary provided by Stockholm's University\footnote{\url{https://teckensprakslexikon.su.se/sok/handalfabetet}}), their data only include one or two examples per sign. This is not sufficient to train an accurate ML model. A solution to the problem of having a an small amount of data is to use a pre-built and pre-trained model and applying transfer learning~\cite{Lia2019}. Transfer learning is a technique that is based on using models trained on one quantitatively large dataset, and then the first layers of this model are reused in order to personalize new models based on a set of limited and specific data.

Several studies on sign language interpretation are based on AI solutions, as well as AI solutions with transfer learning. One study focuses on gesture recognition using a CyberGlove and has received an accuracy of up to 100\%~\cite{Weis1999}. Even though this solution has resulted in a high accuracy, using a CyberGlove device might not be possible at all times and therefore a solution based on using only a camera can be more adaptable. Another study using AI generated a dataset using YouTube videos, but did not use transfer learning~\cite{Quirk2018}. This made their model highly dependent on their dataset and might not generalize well to new situations. A study focusing on using both AI and transfer learning however did not receive higher accuracy than 66\% due to a too small dataset (1200 images)~\cite{Dhi2017}. Based on these studies, generating a larger dataset to use with transfer learning, as well as accurately interpret sings using a single camera, is the focus of this paper. 

Specifically, in this paper we investigate to what extent Convolutional Neural Networks (CNNs) and transfer learning are effective techniques to interpret the hand alphabet of SSL. To find the network architecture with the highest validation accuracy, three different pre-trained models, two optimization algorithms, and three values of the number of frozen layers and step-size, are tested. Our final network structure is based on the pre-trained model InceptionV3, the optimization algorithm of mini-batch gradient with a step-size factor of 1.2, and with five frozen layers. The final model consists of 25 488 698 parameters and 316 layers. Based on this model, we build an application that  translates the signs into its corresponding text form. The final network has a testing accuracy of 85\%. 

In summary, this paper makes the following contributions:

\begin{itemize}
  \item A CNN that translates images from the SSL hand alphabet;
  \item An original approach based on transfer learning, which adapts the model to perform well using a limited training dataset;
  \item An evaluation of the model on $8$ study subjects, obtaining $85\%$ of overall accuracy;
  \item A publicly available implementation of our approach\footnote{\url{https://github.com/gustafvh/SignInterpreterSSL}} and a web application where a static sign is entered as an input image by the user, and its corresponding text form is displayed on the screen \footnote{\url{https://sign-interpreter-ssl.herokuapp.com}}.
\end{itemize}

The rest of this paper is structured as follows. \autoref{sec:background} presents a background of CNNs and transfer learning. \autoref{sec:methodology} describes the methodology, the dataset employed, and the architecture of our CNN model. \autoref{sec:results} presents the evaluation procedure and results obtained when measuring the performance of the model. \autoref{sec:discussion} discusses the challenges associated with learning from images when translating sign language. \autoref{sec:back:related} presents the related work, and \autoref{sec:conclusion} concludes the paper.

\section{Background}\label{sec:background}

Image recognition is the technology of analysing patterns in images in order to classify the image as a particular object~\cite{Shu2020}. An image recognition method generally includes four steps: 1) \textit{image acquisition}: retrieves unprocessed images from a source~\cite{Shu2020} and defines class belonging for each image, these images and the corresponding classes will represent the dataset for the model; 2) \textit{image processing}: performs processing on the image, for example reduces the colour of the background, and finally represents every digital image frame as matrices of pixels; 3) \textit{feature analysis}: is the step where the method chosen comes in at most in analysing the features of the images~\cite{Fuji2019}, and finding patterns; and 4) \textit{image classification}: classifies new, unseen, images to a class among the predefined classes. 

There are several image recognition methods, \eg, statistical pattern recognition~\cite{Shu2020}, fuzzy mathematical method~\cite{Jova2015}, syntactic pattern recognition method~\cite{Shu2020}, and CNNs~\cite{Shu2020}. These three methods have been widely used for image recognition, however, a promising line of methods used today are those that use Deep Learning (DL) and are based on  CNNs~\cite{Fuji2019}. The use of CNNs is particularly suitable for image recognition since they automate the process of feature extraction from the images efficiently. Another advantage of using CNNs, as opposed to the rest of methods mentioned, is the fact that the other methods include feature vectors extracted using algorithms made by researchers. The patterns are thus determined by the researchers and might not actually represent the real nature of the image. CNNs, on the other hand, does this extraction independently of the researcher, and can therefore decrease bias introduced by the researchers by deriving meaning from patterns too complex to be noticed by humans or traditional algorithms~\cite{Ster1996}. CNNs also outperforms other image recognition methods available today~\cite{Fuji2019}.

\subsection{Convolutional Neural Networks (CNNs)}

\begin{figure}
    \centering
    \captionsetup{justification=centering}
    \includegraphics[width=0.4\paperwidth]{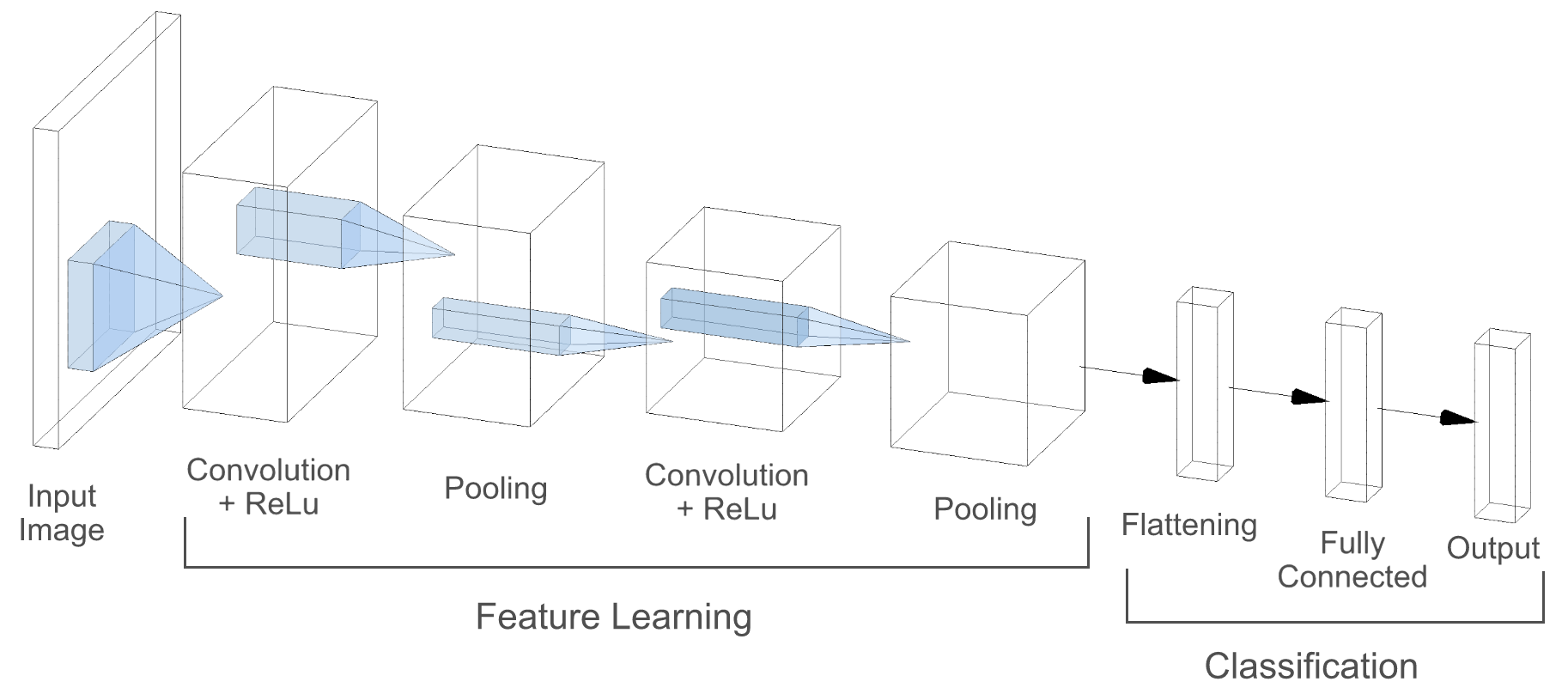}
    \caption{General architecture of a Convolutional Neural Network. Starting from the left in the figure, the input image is split into several smaller matrices that are used as input to the first convolutional layer. This layer performs convolving and the method of Rectified Linear Unit. The next layer performs pooling on matrices. These two steps are repeated several times and comprise the feature learning in the network. The final part, classification, consist of the methods of flattening, fully connection, and finally, providing the output.} 
    \label{fig:image-cnn}
\end{figure}

CNNs is a class of neural networks that are particularly accurate when applied to image recognition tasks. The training of a CNN is performed using the backpropagation method~\cite{Shan2020}. The general structure of a CNNs is described in Figure \ref{fig:image-cnn}. The figure was made with the help of the tool NN-SVG\footnote{\url{http://alexlenail.me/NN-SVG}}. As seen in the figure, CNNs performs feature extraction and classification. Firstly, the input images are split into small matrices of pixels. The feature extraction consist of the two processes of convoluting (and ReLu) and pooling that are applied repeatedly several times~\cite{Fuji2019}. This enables the possibility to recognize specific geometrical patterns, with little data to train on~\cite{Shan2020}. Further on, the classification is performed through the three layer of flattening, full connection, and output. All of these steps are further described below. 

\textbf{Convolutional Layer.}
The first time a input image is entering a convolutional layer, the image is split into many small images in matrix format~\cite{Shan2020}. The convolutional layer consists of a layer of neurons each connected to a filter. A filter is a matrix of weights. The matrices from the input image are assigned to one neuron. All neurons in a layer apply the same filter on these matrices. This allows different neurons to activate on different patterns in the images. The filters perform dot multiplication with its assigned section of the input matrix. All of these scalar values are then stored as representatives for that section in a new matrix. The convolution layer then performs the step of ReLu which increases non-linearity and reduces the unwanted pixels. It replaces all negative values with zero. These values are put in a new matrix which is the new feature map that is transferred to the next layer, the pooling layer~\cite{Fuji2019}.

\textbf{Pooling Layer.}
The output matrix of ReLu is passed onto the pooling layer~\cite{Shan2020}. The pooling layer further reduces the size of the convolved feature map~\cite{Fuji2019}. It takes the input image and divides it into small patches~\cite{Shan2020}. It then takes a specific value from every patch and places it in a new matrix. This specific value can either be the maximum value, minimum value, or average value. This layer downsizes the matrices and thus works as a regularizer for the network, and makes the network focus on main features. This new matrix is the new pooled feature map that is transferred to the next layer~\cite{Fuji2019}.

\textbf{Flattening, Fully Connected Layer, and Output Layer.}
When all the steps of feature learning has taken place, the final pooled feature map needs to be flattened~\cite{Xu2020}. This is performed by transforming the feature map matrix into a one column matrix. The penultimate layer of the network is the fully connected layer~\cite{Shan2020}. It performs a multiplication with the flattened matrix and a weight matrix, and adds a bias vector in order to classify the images into the predefined classes. The final layer, the output layer consist of the probability of class-belonging to each class~\cite{Fuji2019}. However, before outputting the results, the softmax function is used to make sure the probabilities of each class is between zero and one~\cite{Xu2020}.

\subsection{Transfer Learning}
\label{sec:back:tl}

One problem with CNNs is that big networks need high GPU performance which is often difficult to achieve on personal computers~\cite{Ahmad2016}. Without this, the learning will be slow. Another problem is the need for data. A possible solution for this is using a pre-trained model with the technique of transfer learning~\cite{Shen2020}. Transfer learning leverages knowledge from one source to improve learning on another~\cite{Shen2020}. In the following, we describe the general method and the concept of a pre-trained model, together with several models considered for this paper. 

\textbf{General Method.}
\label{sec:back:tl:approach}
Transfer learning firstly uses a pre-trained DL model on a problem~\cite{Lia2019}. It does not have to be based on the same type of input data as the new source~\cite{Shen2020}. However, the performance of the new network will vary depending on which pre-trained model that is used. The first layers from this network are then frozen and put in front of new layers that have not been trained on any data. The learned parameters from the pre-trained source are saved as a vector \(\boldsymbol{\theta} = \big\{\theta_1, \theta_2,..., \theta_n\big\}\) which is transferred to the new model together with new, specific data, for the model to train on. Transfer learning eliminates the need to train the entire network by transferring the knowledge of the pre-trained model and thus reducing the need for large quantities of data.  

\textbf{Pre-Trained Models.}
\label{sec:back:tl:pre-train}
The pre-trained models considered for this paper are all available for use with Keras\footnote{\url{https://keras.io/applications}}, which is the DL API that is used for this paper. All models are specialized in image classification, trained on the dataset ImageNet\footnote{\url{http://www.image-net.org}}.

Table \ref{tab:pre-trained-models} presents the models considered in this paper. The first column shows the name of the model, the second column presents the models' accuracies on the ImageNet dataset. The accuracy is an important indicator to take in consideration when choosing model since the better the model performs on the ImageNet dataset, the better starting conditions for the new model~\cite{Canz2016}. The third column in the table presents the number of parameters representing the model. The higher this number is, the more time and space will it take to train the network. Too many parameters can lead to a slow and memory expensive process on modern computers, however, too few parameters will likely make the pre-trained model less fit for use on new data. 
\input{tables/pre-trained-models}

\subsection{Evaluation Criteria for CNNs}
\label{sec:back:eval}
Evaluating a ML model is about finding the difference between the predicted output by the model, and the actual output~\cite{Hecht1992}. This defines the model's accuracy. Two metrics is tested in the paper to evaluate which one of them performs best on the data. The metric of misclassification error is used to maximize the accuracy. The metric of using a loss function is used to minimize the loss on the data. These two metrics are chosen since they are commonly used by other researchers~\cite{Chris2014, Shen2020, Xu2020, Shan2020}.

The metric of misclassification error is calculated as the average number of correct classifications~\cite{Maki2020}:

\[err(f,D) = \frac{1}{N} \sum_{i=1}^{N} lnd(f(\mathbf{x_i} \neq y_i) \]

where \(lnd(f(\mathbf{x})=1)\) if x is true, otherwise \(lnd(f(\mathbf{x})=0)\). The metric of using a loss function is commonly used on image recognition problems based on a DL network~\cite{Fuji2019}. A distance is here the distance between the feature vectors of the current pattern and the input image. The distance between two images of the same object is considered small, whilst the distance between two images of different objects is considered large. 

The two metrics are evaluated in the beginning of the testing process, before the more specific details of the network is tuned, and the metric producing the highest accuracy on the data is the one used for the rest of the paper. 
\section{Materials and Methods}\label{sec:methodology}

\begin{figure*}
    \centering
    \captionsetup{justification=centering}
    \includegraphics[width=0.65\paperwidth]{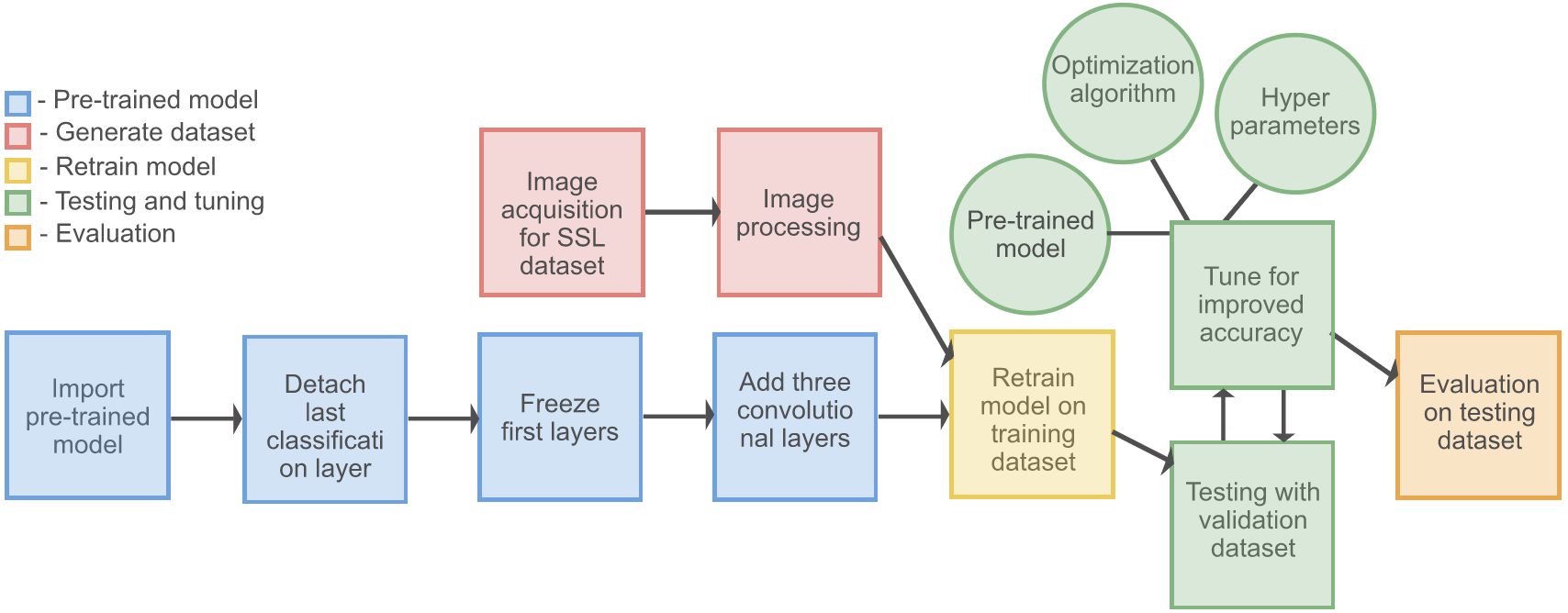}
    \caption{General overview of our approach, including building the network and generating the dataset. The first activities includes changing the imported pre-trained model. Simultaneously the dataset is generated. Following this, the new model based on the pre-trained model is trained using the training dataset. The model is then retrained and the accuracy is improved in several cycles focusing on choosing the combination with the highest validation accuracy. Lastly, the final network is evaluated using the testing dataset.}
    \label{fig:method}
\end{figure*}

The modelling, as presented in Figure \ref{fig:method}, consists of several activities, all of which include quality assurance. The first activity, highlighted in blue, is to handle the pre-trained model. The model is imported and the last classification layer is detached from it. The first layers of the pre-trained model are frozen in order to keep its knowledge. Three convolutional layers are then added at the end of the network. At the same time as this, the SSL dataset is generated, highlighted in red in the figure. This includes steps of image acquisition and image processing. Further on, the model is retrained, yellow color in the figure, on the pre-trained model and the new training dataset. This model is then tested and its accuracy is improved in several cycles, as presented in green in the figure. The focus of the tuning of the network is to find the combination of pre-trained model, optimization algorithm, and tuning of hyper parameters, with the highest resulting validation accuracy. Finally, the final network is evaluated using the testing dataset, highlighted in orange in the figure. The rest of this section is dedicated to the these five phases of the methodology.

\subsection{Pre-Trained Models}
\label{sec:method:apphase:pre-train}
The first step of modelling is to choose and integrate a pre-trained model. A pre-trained model is used as basis of the program to make better predictions on a small dataset. The data for the pre-trained model is already collected, processed, and the model is trained on it; there are thus several variables, such as the type and characteristics of images sued for training the model, that cannot be controlled. This is the part of the data collection that is based on experiments, since the pre-trained models considered all are based on a large dataset. The pre-trained models used in this research are the ones presented in Section \ref{sec:back:tl:pre-train}. They are trained on the ImageNet dataset. 

Since training the models is a very time consuming task: approximately three hours per training session on Colabs GPUs, three out of all of the pre-trained models presented in Table \ref{tab:pre-trained-models} are tested. Table \ref{tab:pre-trained-models} presents eleven pre-trained models that are available via Keras and ordered by their accuracy on the ImageNet Dataset. The three pre-trained models used are decided upon based on the highest accuracy received on the ImageNet dataset, thus InceptionResNetV2, Xception, and InceptionV3 are tested. The process of deciding which one of these three models to use in the final network is presented in Section \ref{sec:method:apphase:imp}. 

The pre-trained models are imported as a pre-existing module through Keras and are then possible to use in the code when for example testing the model on data. The last classification layer is detached from the model and the 20 first layers are frozen in order to keep that knowledge. Finally, three convolutional layers are added to specialize on the new data. The process of choosing and using the pre-trained models is further described in Section \ref{sec:method:apphase:imp}.

\subsection{Swedish Sign Language Dataset}
\label{sec:method:apphase:data}
The next step is to collect and process the data of SSL with a focus on the  generation of a generic dataset. Image acquisition and image processing, are conducted in this step of the process. Feature analysis and image classification, are performed in a subsequent step, when the model is retrained and tested on the new dataset. 

\subsubsection{Image Acquisition}
\begin{figure}
    \centering
    \captionsetup{justification=centering}
    \includegraphics[width=0.20\paperwidth]{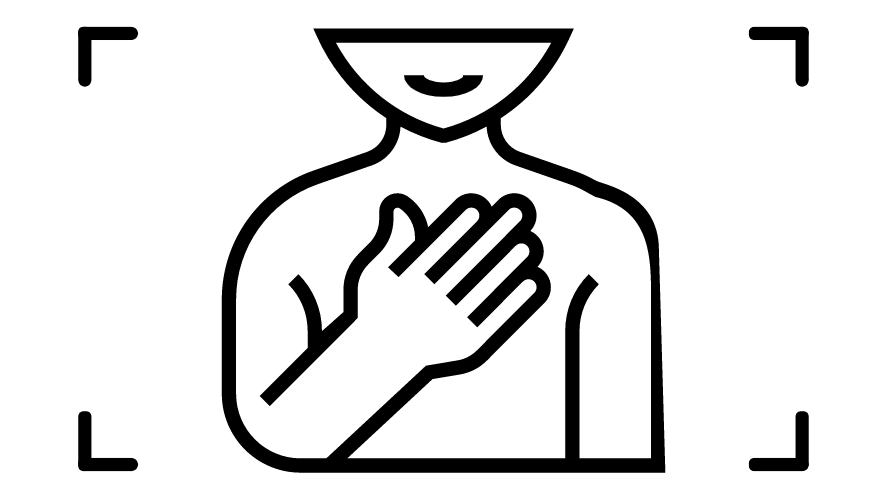} 
    \caption{Illustration representing how the signing videos are staged regarding the position of the person's body and hand.}
    \label{fig:design-body-sign}
\end{figure}

The first step, image acquisition, is performed by filming all the letters in the hand alphabet during 20 seconds, for a total of eight times by five subjects. All 26 letters are recorded eight times, and every recording is performed signing with the person's right hand. The recordings are performed as presented in Figure \ref{fig:design-body-sign}. The figure is created using assets from Flat Icon\footnote{Figure \ref{fig:design-body-sign} uses assets from the icon made by Eucalyp from \url{https://www.flaticon.com}}. Only a part of the body and face are included, the person is centered, and their hand is centered in front of the chest. The webcam used is the built-in FaceTime HD camera from Apple Macbooks, which is rated at 1.2 megapixels and records 720p video in 30 or 24 frames per second. 

The eight recordings are the basis of the training, validation, and testing datasets. The signs are based on the SSL dictionary provided by Stockholm's University. The conditions between the recordings are varied by background and clothes. When filming, the hands are slightly rotated to allow for more variance. Each sign of a letter is then classified as belonging to that specific letter.

The datasets are  collected by five different subjects. This part of the method connects to the consciousness of ethics in society and technology. This is one part of the investigation where the technology could become an inhibitor and not an enabler for sustainable development. By collecting diverse data, it could improve the ML models ability to generalise and different types of signers and hands. 

\subsubsection{Image Processing}
The second step, image processing, is now performed on each video. A function is created to read all videos per letter and create one frame per 0.1 second so every recording becomes 200 images. These images are then resized to a size of \(224 \times 224\) pixels. 100 of these images are placed in the training set, 50 in the validation set, and 50 in the testing set. All images are put into repositories on GitHub, cloned into the code and then merged once on the Colab Virtual Machine in order to be used as data\footnote{\url{https://colab.research.google.com/drive/1oxKUHDykfQOsG0VpP_pblMf9waEy8pq7?usp=sharing}}. To prevent memory overflow, the images in the training dataset are split up into batch sizes of 32 that are sent into the model in batches. 

The training images are then augmented in order to allow for more noise in the data. The testing images are also augmented in order to create a more realistic real-life testing experience. The steps performed for both datasets are rotation, shifting height and width, zooming, and tilting, to allow for more variations of signing. When shifting and tilting, the points now outside of the boundaries of the image are filled with the nearest pixel. The testing images are also altered on brightness. No image augmentation is performed on the validation dataset. The image augmentation is performed with the Keras class ImageDataGenerator \footnote{\url{https://keras.io/api/preprocessing/image/\#image-data-preprocessing}}. 

\subsection{Model Retraining}
\label{sec:method:apphase:train}
When the pre-trained model is imported and the new dataset generated, the training of the new model begins. This is done by adding extra layers. Three convolutional layer are added. The first two consist of 1024 nodes each, the third consist of 512 nodes, and they all use ReLu. These values of the numbers of nodes are chosen since they are frequently used in other studies and yielded satisfying results. The pooling layers use global average pooling. Finally, softmax is used to ensure that the probabilities end up between zero and one. The first 20 layers of the pre-trained model are frozen (model weights became immutable) in order to not retrain them, but keep their knowledge. Then the model is trained on the new data with the new layers in order to specialize on the new data. This is the step of feature analysis in the image recognition framework. The final step of image classification then occurs when testing the model and finally towards the end of the implementation, when using the application. 

\subsection{Model Testing}
\label{sec:method:apphase:test}
Before starting to improve the accuracy of the model, the evaluation method needs to be decided upon. This focused on the research's approach, deductive, which ends with specific accuracies of the model's performance. As the model used is a neural network, the outcome is a generalisation based on the collected data over several runs explaining the results as relationships between several variables.

The model is trained on 50\% of the data, validated on 25\%, and tested on 25\%. This split of the data is done on a per person and sign-level, meaning the training, validation and test datasets each consists of an even distribution of each sign and person. The testing data therefore contains samples from every person and sign and consists of 25\% of the full dataset. The evaluation of accuracy on the validation data is  performed both through the loss function and the total misclassification error on the validation data. The models are non-deterministic and thus each run of the model generates a slightly different accuracy. Each run is repeated three times in order to get an average accuracy. Several training sessions on ten epochs, with different parameters, are tested, as will be presented in Section \ref{sec:method:apphase:imp}. These runs are used when comparing different networks and thus the differences between the networks are of importance, not its absolute performance. In order to present the final accuracy of the model, the final network architecture is trained during 30 epochs. This run is then tested on the testing data to present the final accuracy. 

A dynamic video tool is developed to visualize the model's performance on single signs. This is also developed to be able to detect several signs in a row and help evaluate the different models accuracy and behaviour in real-life situations and with fast feedback. It works by making predictions continuously on every frame from the webcam video stream and showing the result live. Basic helper functions including backspace, delete, and reset, are also added. The tool is only developed to be run locally for testing purposes as deploying that functionality on the web is out of scope for this investigation.

\subsection{Accuracy Improvement}
\label{sec:method:apphase:imp}
One part of the research that focused on the experimental aspect of our methodology is the choice of the  pre-trained model to use, the optimization algorithms, and the tuning of hyper parameters. The methodology for improving the model's accuracy is conducted in two steps. Firstly, the InceptionResNetV2, Xception,and InceptionV3 pre-trained models are tested with different optimization algorithms. Secondly, two hyper parameters are tuned based on the combination of pre-trained model and optimization algorithm that has the highest accuracy form the previous step. In this section, we give details regarding those improvements.

\subsubsection{Model Pre-training and Optimization}
The three pre-trained models are then tested on two optimization algorithms. Due to the same reasoning as above, not all algorithms are tested. The two methods tested are mini-batch gradient descent and Adam since they are commonly used by other researchers. Thus, all three pre-trained models are tested on the two optimization algorithms. The tests are conducted using a fixed batch size for the optimization algorithm of 32, and on ten epochs. 
Based on these tests, the architecture with the highest accuracy is chosen, according to the Wilcoxon signed rank test~\cite{Liu2018}. This test is appropriate to use since it can determine, with statistical significance, if one model combination can be used over another based on its overall performance. The combinations of different pre-trained models are used in the tests to determine if the models perform as an equal distribution or not and thus which validation accuracies are the most reliant. The final combination with most statistical significance is chosen as the final combination. With the final combination determined, the tuning of the rest the parameters is performed. 

\subsubsection{Hyper Parameters Tuning}
The tuning of hyper parameters begins with one combination of pre-trained model and optimization algorithm. This is done with the help of the tool Weights \& Biases \cite{wandb}. It allows sweeping of several parameters in order to help find the parameter values that maximize the networks performance. The following two parameters are tuned: the number of frozen layers of the pre-trained model, and the step size factor. The step size factor is a factor which determines the relationship between the total number of training data examples. The number of frozen layers determines the impact the pre-trained model should have on the final outcome. The step size factor is tested on three values, namely 0.2, 0.7, and 1.2. Furthermore, the number of frozen layers is tested on three values, namely, 5, 20, and 50.

\begin{figure}
    \captionsetup{justification=centering}
    \includegraphics[width=0.41\paperwidth]{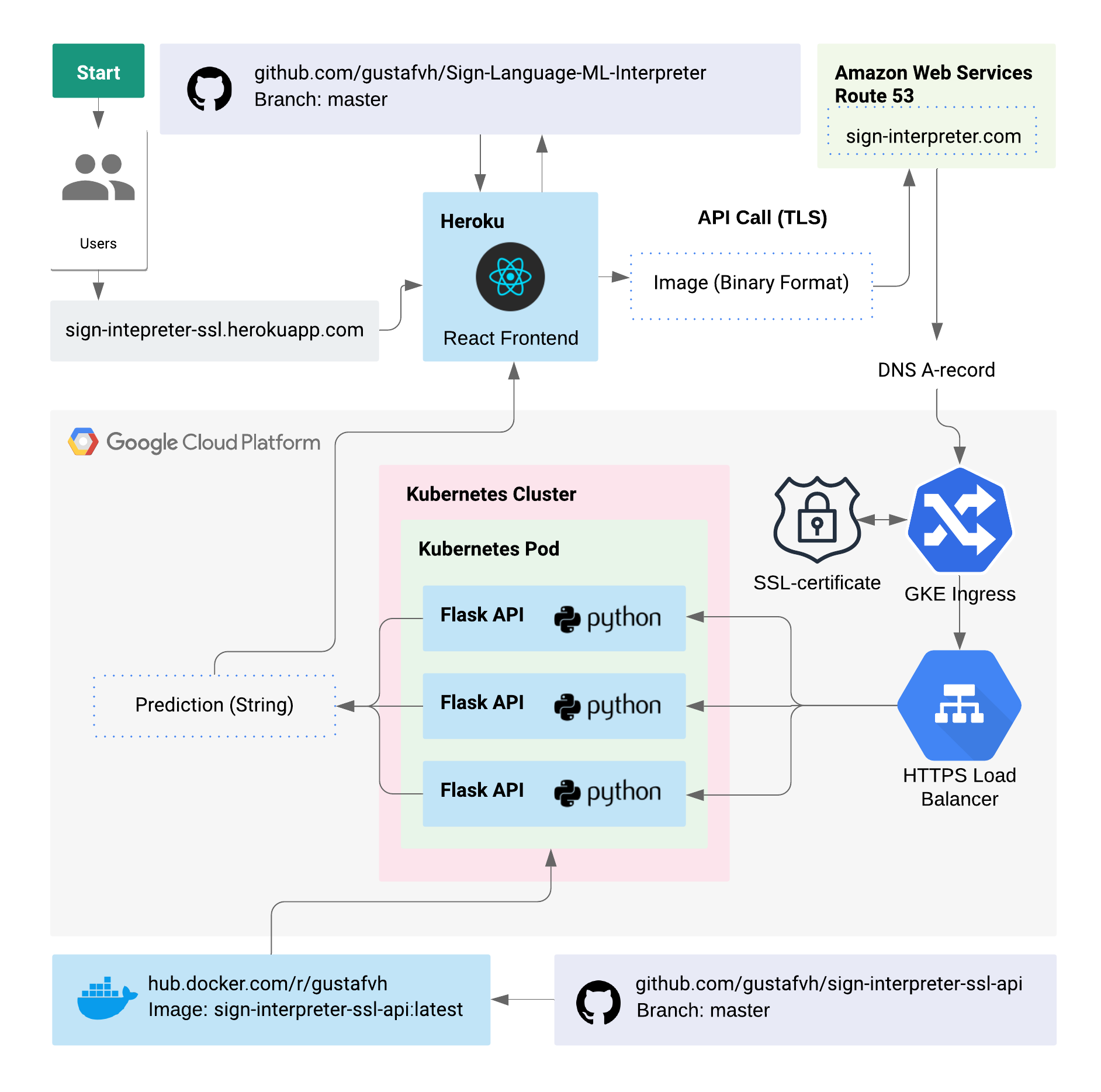} 
    \caption{Cloud architecture of the web application. The arrows represent flow of data for an API request when a user wants to analyse a sign. Starting at the top left corner, the user enters the URL in a browser to the Heroku-hosted application which pulls its source code from GitHub. The front end (client) makes a POST request with the sign image now binary encoded, to the sign-interpreter.com domain, which points to the cloud services. That entry point is an ingress that validates the identity with a certificate to be able to handle the HTTPS-protocol. The request is then forwarded to the load-balancer which sends it to one of three Flask Python back ends that return a prediction to be returned to the client and displayed in the application. The Python back ends are based on Docker images living on Docker Hub and GitHub.}
    \label{fig:cloud-architecture}
\end{figure}

\begin{figure}
    \centering
    \captionsetup{justification=centering}
    \includegraphics[width=0.195\paperwidth]{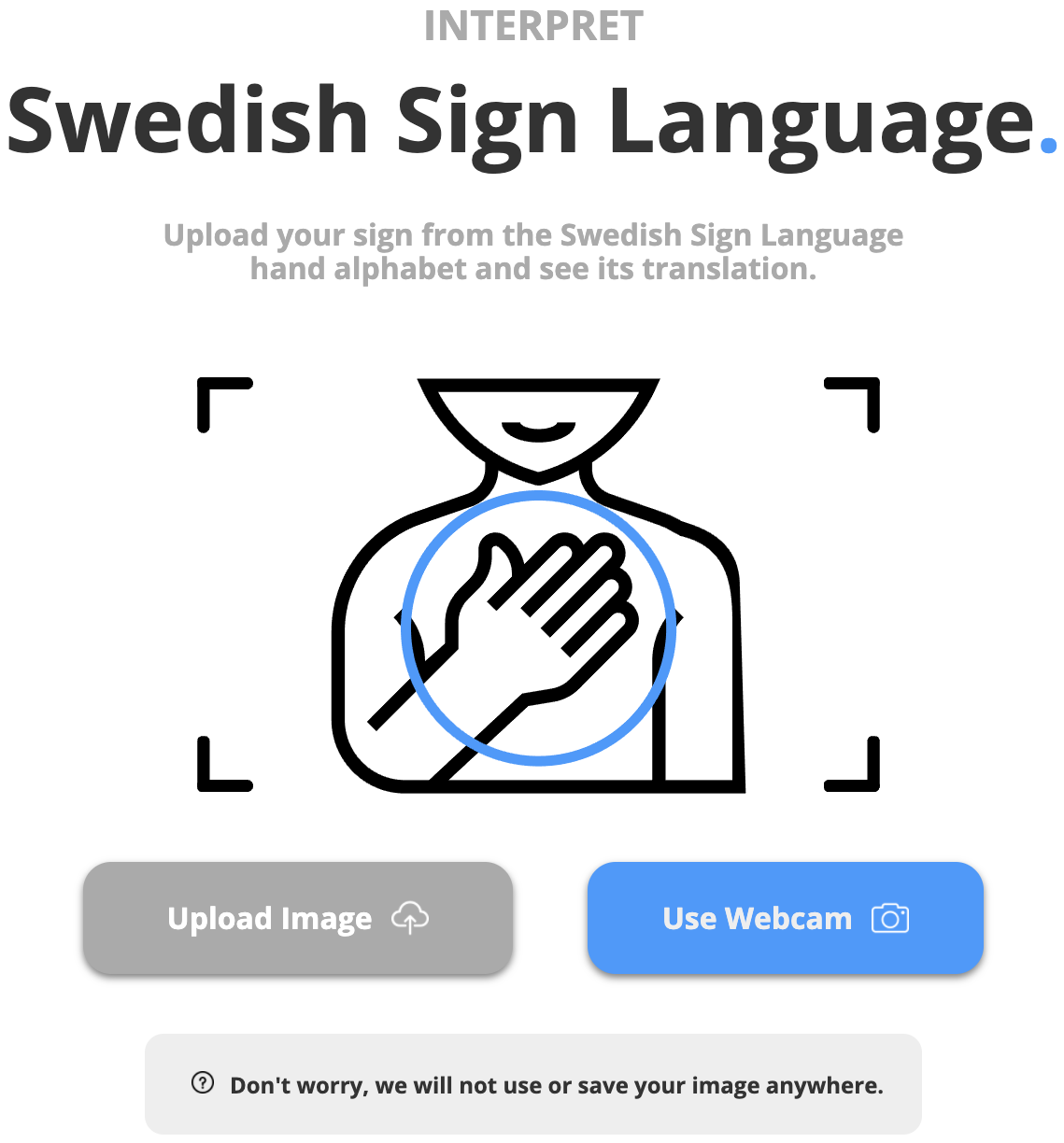} 
    \includegraphics[width=0.195\paperwidth]{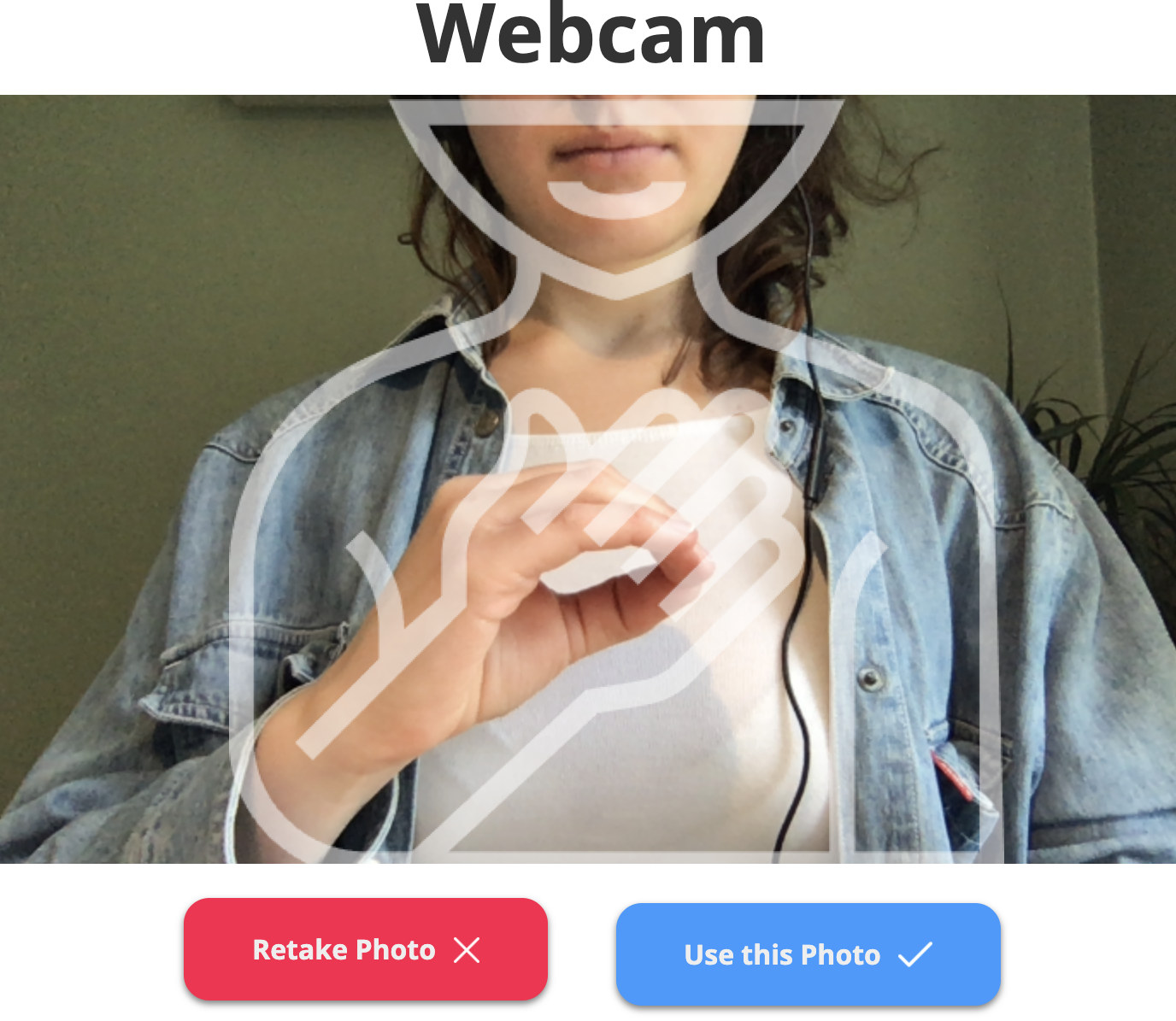}
    \includegraphics[width=0.4\paperwidth]{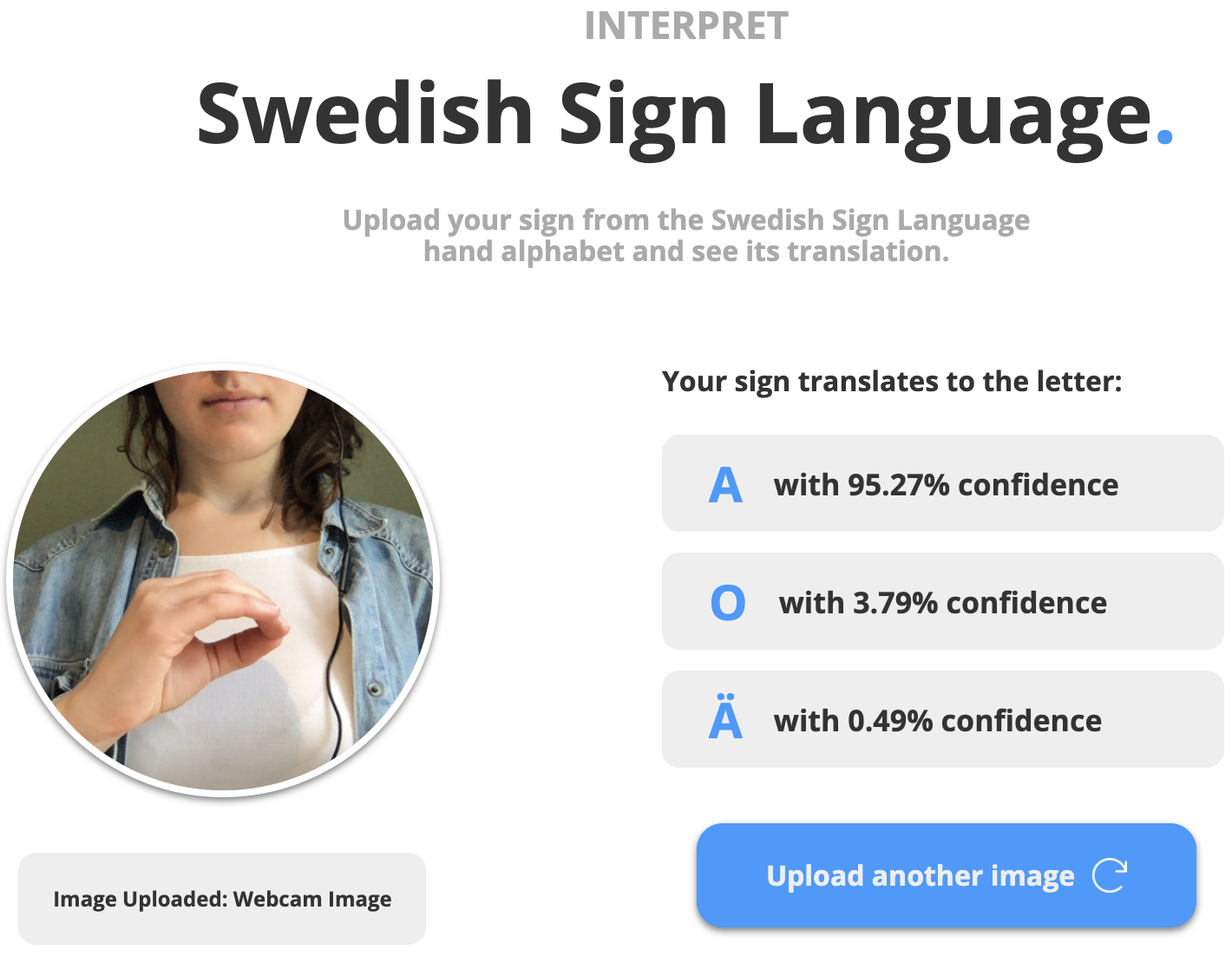}
    \caption{User flow of the web application when using the webcam to take a picture. The first image presents the home page where a button for using your webcam is placed on the bottom right. The second image presents the page of the webcam where the user can take, and re-take, a picture. The final image presents the results page that shows the sign with highest confidence on the top.}
    \label{fig:app-steps-webcam}
\end{figure}

\begin{figure*}
    \centering
    \captionsetup{justification=centering}
    \includegraphics[width=0.12\paperwidth]{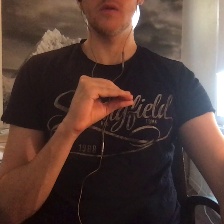}
    \includegraphics[width=0.12\paperwidth]{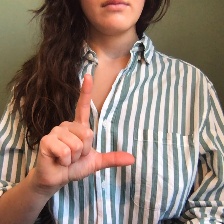}
    \includegraphics[width=0.12\paperwidth]{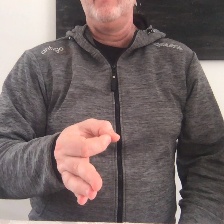}
    \includegraphics[width=0.12\paperwidth]{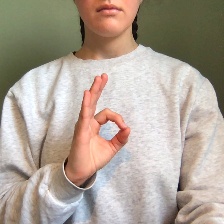}\\
    \includegraphics[width=0.12\paperwidth]{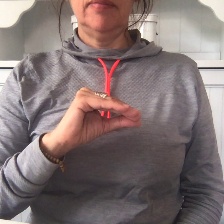}
    \includegraphics[width=0.12\paperwidth]{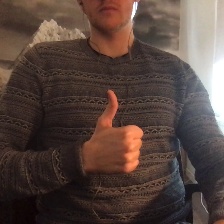}
    \includegraphics[width=0.12\paperwidth]{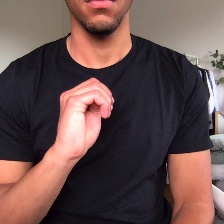}
    \includegraphics[width=0.12\paperwidth]{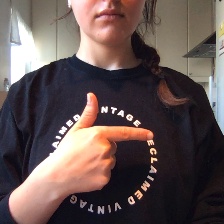}
    \caption{One image per recording in the dataset for SSL. The images are showed in the same order as the recordings shown in Table~\ref{tab:data-experiments}. The images represent the letters A, L, P, H, A, B, E, and T.}
    \label{fig:signs-SSL}
\end{figure*}

\section{Experimental Results}\label{sec:results}

\input{tables/data-experiments}

The goal of the experiments is tuning a network with the highest resulting validation accuracy, which allows the network to interpret as many words as possible correctly. The objective of each experiment is thus to find the combination of parameters that provide the highest accuracy.

All code and data for this investigation is stored in three different GitHub repositories where each repository contains a \texttt{README} file with more information of its content. The code for the model, data generator, and front end for the application, can be found in one GitHub repository \footnote{The following link contains the GitHub repository for the model, data generator and front end for the application: \url{https://github.com/gustafvh/SignInterpreterSSL}}. The code for the back end API for the application is stored separately and can be found in another GitHub repository \footnote{The following link contains the GitHub repository for the back end API code: \url{https://github.com/gustafvh/SignInterpreterSSL_API}}.

Our dataset is collected using five different subjects, in a total of eight recordings. Figure \ref{fig:signs-SSL} presents one representative image per person, or recording, that is included in the dataset. More information about each recording is shown in Table \ref{tab:data-experiments}. Two important factors that might change how an image of a signer looks visually is age and gender which is why it is important to have a varied selection of this in the dataset. This is to ensure the model becomes better at generalising and as signer-independent as possible. As previously mentioned, the exact number of images each person contributes with, depends on how long each recording of each sign is and therefore varies slightly.  

The testing data is generated from the recordings of these persons (but never overlaps the training or validation data) and is later further augmented to make them even more significantly different to the training and validation-data. These image augmentations include rotations, width and height shifts, shearing, zoom and brightness alterations to ensure the model is fairly evaluated using new and previously unseen data.

\subsection{Accuracy Testing}
\label{sec:results:acc}
This section focuses on the tests conducted to find the tuning of the network with the highest resulting accuracy. This is performed in two steps, first with three different pre-trained models and two different optimization algorithms, presented in Section \ref{sec:results:acc:opt-train}, and then by tuning the hyper parameters, presented in Section \ref{sec:results:acc:hyperparam}. 

\subsubsection{Optimization Algorithm and Pre-Trained Model}
\label{sec:results:acc:opt-train}
These first tests are conducted using two different metrics as stated in \autoref{sec:back:eval}. Firstly, the parameter for using a loss function is activated, aiming to minimize the validation loss. Secondly, the metric of misclassification error is used, aiming at maximize the validation accuracy. Minimizing the validation loss results on average in 20\% lower accuracy on the data. Therefore, maximizing the validation accuracy is used for the rest of the investigation.

The results of the tests of the pre-trained models and optimization algorithms are presented in Table \ref{tab:accuracy-results}. The table shows that the differences in accuracy between the different combinations are small. Thus, the Wilcoxon signed rank test is used to validate which combination proves to perform best on the data.

\input{tables/accuracy-results}

\subsubsection{Statistical Validation}
Since the runs with InceptionResNetV2 produces significantly lower accuracies than the other pre-trained models, they are excluded from the Wilcoxon test. However, Xception and InceptionV3 produces similar results and thus the test is used to see if there are any statistical significance in the data that could be used to determine which combination performs better. 

The first test keeps Xception static while altering the two optimization algorithms. The second test keeps InceptionV3 static while altering the two optimization algorithms. The data used for the tests are the validation accuracies received per epoch of training for the different combinations. The null hypotheses for both tests are that the combinations are of the same distribution.

The p-value of the test with InceptionV3 is approximately 0.0069 and thus the null hypothesis can be rejected at a confidence level of over 99\%. This meant that with InceptionV3, the optimization algorithms are not of the same distribution with a certainty of over 99\%. This means that the combination that provides the highest accuracy can be used. Since mini-batch gradient descent produces a higher accuracy than Adam for InceptionV3 as shown in Table \ref{tab:accuracy-results}, it is passed on to the next step of the evaluation. Approximately the same reasoning can be used regarding the test with Xception since it produces a p-value of approximately 0.0469. Its null hypothesis can also be rejected, however here at a confidence level of over 95\%. Since Adam performs a higher accuracy than mini-batch gradient descent as shown in Table \ref{tab:accuracy-results}, it is passed on to the next step of the evaluation. 

The two final combinations to evaluate is thus InceptionV3 with mini-batch gradient descent, and Xception with Adam. As seen in Table \ref{tab:accuracy-results}, they produce a validation accuracy of 94.78\% and 94.77\% respectively. Thus, further tests are needed to show which performs better on the data. To evaluate this, a box plot is used to show the variance and consistency in validation accuracies for the different epochs.

\begin{figure}
    \centering
    \captionsetup{justification=centering}
    \includegraphics[width=0.4\paperwidth]{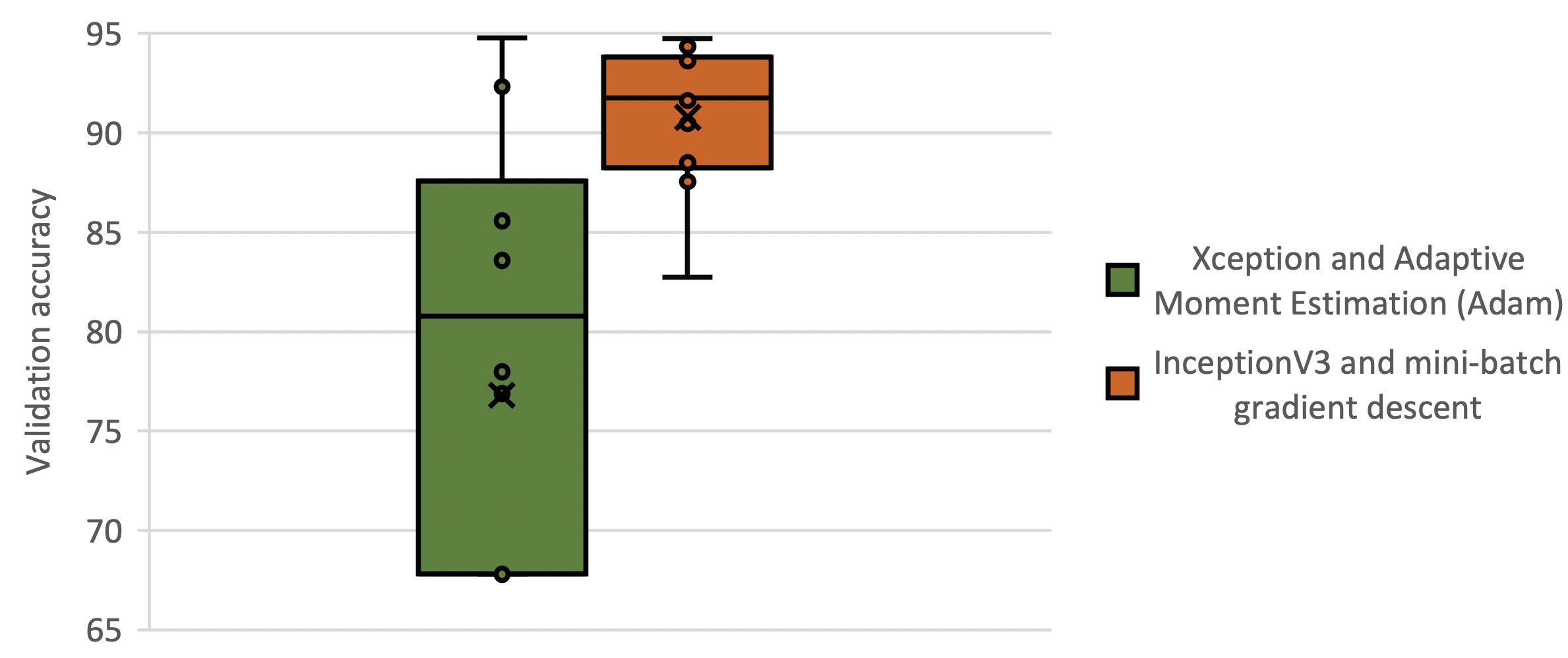} 
    \caption{Box plot of the validation accuracies of the two combinations, Xception and Adaptive Moment Estimation (Adam), as well as InceptionV3 and mini-batch gradient descent, varied over the ten epochs. The boxes include 50\% of the data points and thus shows the main distribution. The lines over and under the boxes represent the highest and lowest validation accuracies received for the different combinations.}
    \label{fig:box-plot}
\end{figure}

\subsubsection{Evaluation of Spread and Consistency of Validation Accuracy}
As seen in Figure \ref{fig:box-plot}, InceptionV3 with mini-batch gradient descent has a low spread in the validation accuracies per epoch and produces consistently high accuracies. Xception with Adam however, shows a large variance in the results as well as consistently lower accuracies than the other combination. This shows that Adam is more prone to get stuck in local minima. Thus, even though the two combinations performs equally as stated in Table \ref{tab:accuracy-results}, InceptionV3 and mini-batch gradient descent is shown to be more reliant since the validation accuracies per epoch are more consistent. This increased reliance is an advantage since it simplifies the tuning of hyper parameters as every run is more prone to deliver good results. This combination also increases the replicability of the experiments. Thus, the combination of InceptionV3 and mini-batch gradient descent is used as the basis for the final network architecture.

\subsection{Hyper Parameters Tuning}
\label{sec:results:acc:hyperparam}
\begin{figure}
    \centering
    \captionsetup{justification=centering}
    \includegraphics[width=0.4\paperwidth]{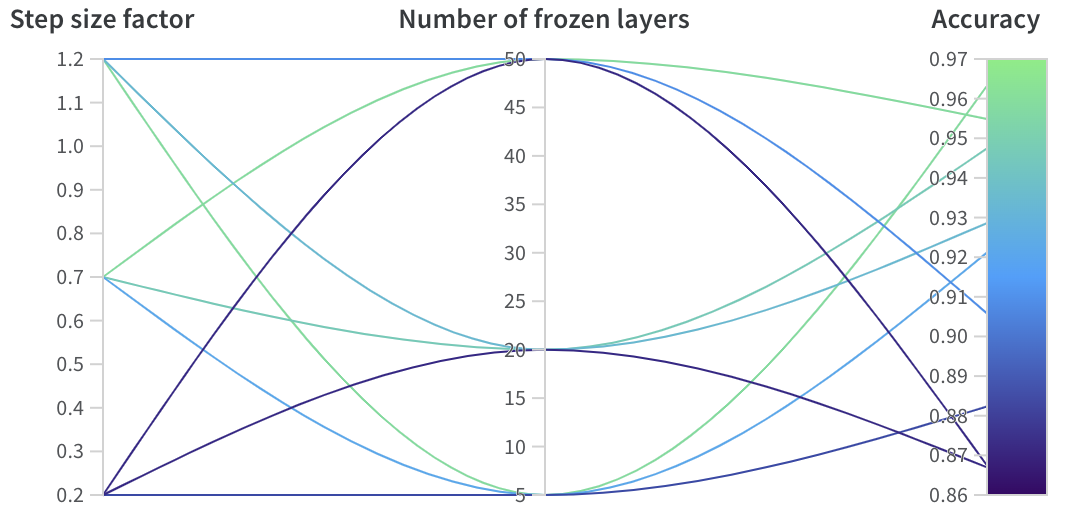} 
    \caption{Results from the tuning of hyper parameters based on InceptionV3 and mini-batch gradient descent. Two hyper parameters are tuned, the step size factor, and the number of frozen layers of the pre-trained model. The graph includes nine different combinations of the hyper parameters and presents the combinations' validation accuracies after training. The figure is made automatically by the tool Weights \& Biases~\cite{wandb}.}
    \label{fig:hyperparam}
\end{figure}

The hyper parameters for the network based on InceptionV3 and mini-batch gradient descent are now tuned. Two hyper parameters are in focus; the step size factor and the number of frozen layers of the pre-trained model. The results from these tests are presented in Figure \ref{fig:hyperparam}. The figure shows that the two combinations with the highest validation accuracy are with step size factor 1.2 and five frozen layers, and with step size factor 0.7 and 50 frozen layers. These two combinations are trained two times more each in order to find the average accuracy over several runs. This results in an accuracy of 96.46\% for the combination of a step size factor of 1.2 and five frozen layers, and an accuracy of 93.90\% for the combination of a step size factor of 0.7 and 50 frozen layers. Thus the final values of the hyper parameters are five frozen layers and a step-size factor of 1.2. The value of the step size factor, corresponds to a learning rate close to, but bigger than, the number of training data samples divided by the batch size.

\subsection{Final Network Architecture}
\label{sec:results:final-arch}
The accuracy tests ends with one final network architecture with the highest validation accuracy. It consists of the following properties: the pre-trained model is InceptionV3, the optimization algorithm is mini-batch gradient descent, the batch size is 32, the step-size factor is 1.2, and the number of frozen layers of the pre-trained model is five. 
To come up with the final architecture, several decisions are made. More specifically decisions regarding which pre-trained model, optimisation algorithms and hyper-parameters to be tested and evaluated. Firstly, only three pre-trained models are tested, based on the highest accuracy, as presented in Table \ref{tab:pre-trained-models}. Then, mini-batch gradient descent and Adam are chosen as leads on optimization algorithms. Mini-batch gradient descent is chosen since it is a well-balanced combination of the basic approaches of Stochastic Gradient Descent and batch gradient descent and is more stable than the others to converge to the global minimum. Furthermore, Adam is chosen since it generally converges fast and is well suited for problems with a large amount of parameters. The combination of InceptionV3 and mini-batch gradient descent is then chosen since it produces the most reliable results based on the Wilcoxon signed rank test as presented in Section \ref{sec:results:acc:opt-train}. Three values of step-size factor and number of frozen layers are tested. These numbers are based on the recommendation from Keras, and then one smaller number and one bigger. The hyper parameter values of a step-size factor of 1.2 and five frozen layers are chosen since they performed better than other values on the combination of pre-trained model and optimization algorithm as presented in Section \ref{sec:results:acc:hyperparam}. The final network architecture is thus based on five frozen layers of InceptionV3. Thus, transfer learning is used, however not to the extent that the network become too specialized on the pre-trained model and its data.

On top of the pre-trained model, three convolutional layer are added. The first two with 1024 nodes, the third with 512 nodes, and they all use ReLu as activation function. The pooling layers use global average pooling. Finally, softmax is used to ensure that the probabilities end up between zero and one. In order to present the final accuracy of the network, it is trained one last time, this time during 30 epochs. The final model, as it is a CNN, consists of 25 488 698 parameters and 316 layers\footnote{All layers and parameters for the network can be found at \url{https://app.wandb.ai/sign-interpretor/sign-interpreter/runs/y11of78x/files/model-best.h5}}. 

\begin{figure}
    \centering
    \captionsetup{justification=centering}
    \includegraphics[width=0.35\paperwidth]{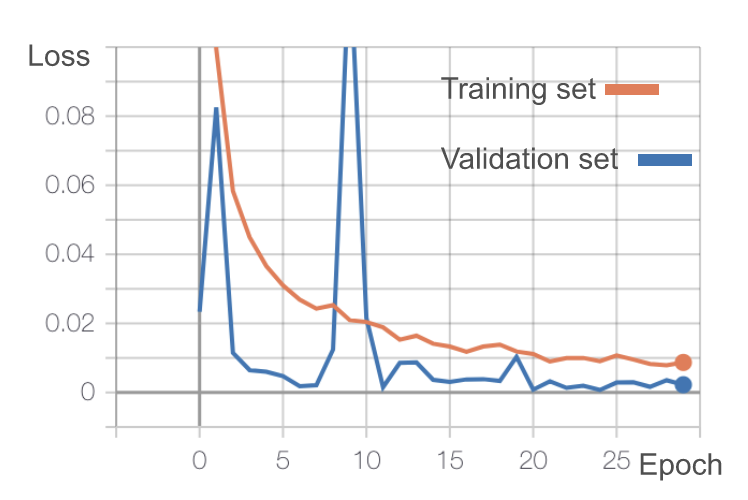} 
    \includegraphics[width=0.35\paperwidth]{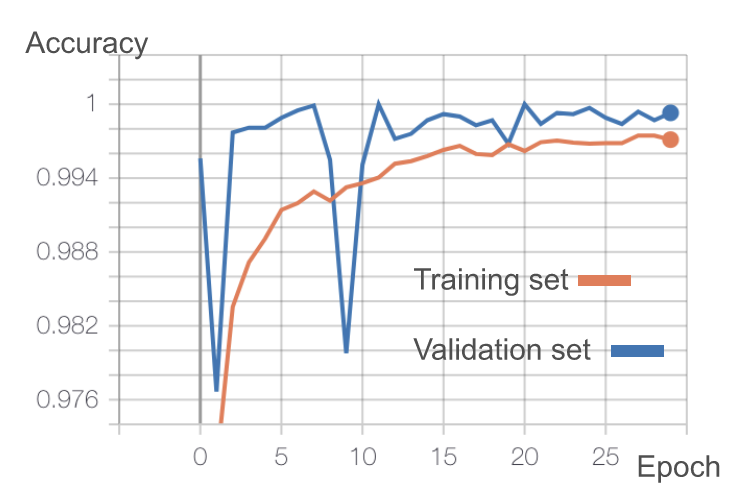} 
    \caption{Loss and accuracy of the training and validation set per epoch for the final model. The first figure presents the loss, and the second figure presents the accuracy on the training and validation set. The figures are made automatically by the Tensorboard tool in the tool Weights \& Biases~\cite{wandb}.}
    \label{fig:loss-accuracy-final}
\end{figure}

Figure \ref{fig:loss-accuracy-final}  shows the training and validation loss and accuracy over the 30 epochs. As we observe in the figure, the training and validation loss soften after approximately twelve epochs. The overfitting that appears in epoch two and nine is eliminated after more epochs. The final model is then tested on the testing dataset. The final testing accuracy is 85\%.

\begin{figure}
    \centering
    \captionsetup{justification=centering}
    \includegraphics[width=0.20
    \paperwidth]{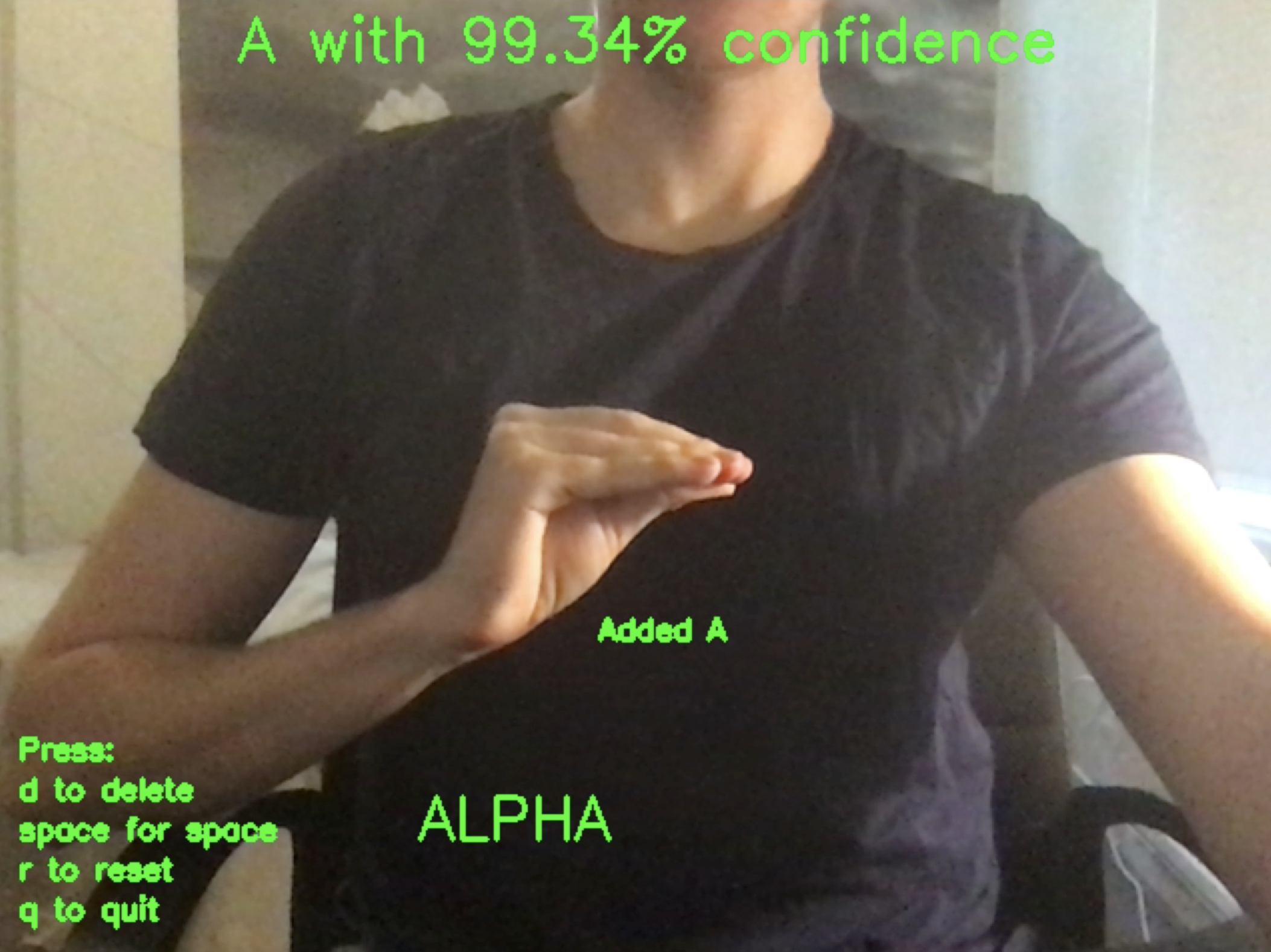}
    \caption{A caption of the video tool used for testing. This picture presents how a list of letters were signed and added to a word. The bottom left of the picture shows the basic helper functions included to be able to sign words.}
    \label{fig:video-alpha}
\end{figure}

The dynamic video tool that is used when testing the models' performances can be seen in Figure \ref{fig:video-alpha}. This particular picture is taken based on the final model. 

\section{Discussion}\label{sec:discussion}

The delimitations can be constricted to those regarding the sign language, and the model interpreting the signs. 

\subsubsection{Sign Language Delimitations}
One delimitation regarding the sign language is that only the static signs of the SSL hand alphabet is used. Words, and four of the letters, in SSL are dynamic. They consist of several signs conducted in a specific order after each other. The remaining 25 letters in the hand alphabet consist of only one still hand gesture, and are thus static. The four letters that are dynamic are: Y, Å, Ä, and Ö. Å is for example the same sign as A, but moved around in a full circle. Any individual frame extracted from the sign Å would be interpreted as A. Y, is a sign shaped like a boat moved vertically down. Even though Y is dynamic, it can be interpreted statically since no other letter is shaped like a boat causing ambiguities, and Y is therefore included in the paper. Å, Ä, and Ö, on the other hand, will not be included in the paper. 

When interpreting words and sentences, facial expressions are often involved when signing and this is not of focus in this paper. The focus is only on hand gestures which works well for the hand alphabet. Another factor when signing is that the room and objects around the signer is often used as reference and commonly used to point at. This is not a critical factor for interpreting the hand alphabet and thus these references are not taken into consideration in this paper.

\subsubsection{Model Delimitations}
One delimitation regarding the model used to interpret the signs is the use of transfer learning. This paper focuses on using transfer learning for interpreting the hand alphabet of SSL, thus it does not necessarily suggest that the same thing can be done for the rest of the sign language, or for the hand alphabets of other sign languages. Furthermore, the paper focuses on using particularly transfer learning, and thus other models with or without the basis of ML will not be tested. Finally, the dataset used for SSL is specifically developed for this paper based on us signing, and thus other possible data sources will not be used. 

\section{Related Work} \label{sec:back:related}

There have been several studies aimed at interpreting sign languages. Some of them have been based on gloves that can interpret signs, while others rely on computer vision and statistical comparisons. Furthermore, there are studies focusing on both SSL and CNNs. This section will present some of these studies. Section \ref{sec:back:related:no} focuses on studies performed without neural networks, while Section \ref{sec:back:related:nn} focuses on studies with neural networks. Finally, Section \ref{sec:back:related:tl} presents studies on sign language and transfer learning. 

\subsection{Sign Language Interpretation Using Human Supervision}
\label{sec:back:related:no}

Hern et al.~\cite{Hern2004} study the signer wearing an AcceleGlove when signing. In this work, the different angles per finger are used as input to a computer program that analyses the positions. It is tested on different subjects and is able to recognize 30 one-handed signs with an accuracy of 98\%. 

Glenn et al.~\cite{Glenn2005} is based on computer vision and statistical comparisons. Each sign is filmed in a controlled environment, the background is extracted, the image is cropped, resized and edge detected, and finally placed in an adaptive statistical database. To classify a sign as a particular word, the image is processed and then compared to all images in the statistical database.

Akram et al.~\cite{Akram2012} used a Kinect sensor to recognise SSL signs. The signer uses a RGB-D Kinect sensor placed in the hand. This allows the backgrounds to be removed, helps with the resolution when the hand was placed in front of the face, and simplifies the use of 3D signs. The classification is done through a statistical database. The results are different depending on who signs the signs. One signer received an accuracy of 77\% while one had 94\%. 

On the other hand, the work of Segundo et al.~\cite{Segundo2008} is conducted with the opposite goal to this paper, to translate spoken words into sign-language. The system uses speech recognition, a natural language translator, and a 3D avatar to show translations for Spanish Sign Language. It achieves a 31.6\% Sign Error Rate. This significantly differs from this paper as one of the most challenging aspects of this paper is to correctly identify the sign from an image, which is not needed during this direction of translation.   

Most of the related work presented includes several forms of assisting equipment, such as depth-cameras and gloves, to simplify isolation and identification of the hands performing the signs. In this paper, the focus is on making an accurate interpretation possible without any specialized equipment besides a basic RGB-camera. 

\subsection{Sign Language Interpretation Using CNNs}
\label{sec:back:related:nn}

The studies presented in this section are based on ANN. The benefits of using neural networks relies on their ability to derive meaning from patterns too complex to be noticed by humans or traditional algorithms~\cite{Ster1996}.

Weis et al.~\cite{Weis1999} focuses on gesture recognition. The experiments are conducted with a CyberGlove, a glove with virtual reality sensors. The analysis is conducted by a multi layered ANN, and the accuracy is close to 100\% for some gestures. Pugeault et al.~\cite{Pugeault2011} present an interactive finger-spelling graphical user interface for ASL, which shows good performance and robustness for multiple users.

Koller et al.~\cite{Koller2016} exploited the discriminitive power of CNNs with application to hand shape classification in the scope of sign language. Moc et al.~\cite{Moc2017} perform a study that aims at recognizing a stream of continuous signs in a video. The computer architecture used is RNNs. The network has feedback connections which is suitable for video processing. The paper reaches an accuracy of 80\% on a continuous stream of video data. 

Quirk et al.~\cite{Quirk2018} translate signs filmed with a webcam. The study aims to translate the Auslan Sign Language alphabet. The dataset for the Auslan alphabet is generated by extracting signs from YouTube videos and drawing boxes over the hands. They use CNNs and the final accuracy is 86\%. This paper can utilise the fact that there were sign language instructional videos available on YouTube for the Auslan Alphabet and thus be used as data, which is something that is not currently available for SSL. Therefore, they do not utilise transfer learning to be less dependent on large datasets.

Cam et al.~\cite{Cam2018} focuses on translating a continuous stream of sign language sentences. Specifically, they improve the interpretations when it comes to grammar. The paper is built on CNNs and attention based encoders and results in several sentences being correctly interpreted. The paper focuses more on aspects of grammar, something this paper chooses to have as a delimitation.  

\subsection{Sign Language Interpretation Using Transfer Learning}
\label{sec:back:related:tl}

The studies presented in this section are based on ANN and transfer learning. The datasets used on the pre-trained models have all been limited.

Moci et al.~\cite{Moci2018} build a system to interpret British Sign Language. The datasets of British Sign Language used are a corpus for standard English with transcriptions to sign language and a pre-processed corpus called Penn Treebank. The videos from the corpuses are split by sentences. The ML architectures used are both based on RNNs and CNNs. The study shows good results in words, but sentences are not interpreted grammatically correct. This study can also utilise the fact that a British corpus existed with more than just a few examples per sign (something not available for SSL). This eliminates the need to create new data for the transfer learning process.

Dhi et al.~\cite{Dhi2017} use transfer learning to interpret Indian Sign Language. A 3D-camera is used to help with depth interpretation, which differs from this paper as it aims to only use a basic RGB-webcam without that assistance. The data used is based both on images per sign but also depth images, which reduces the pre-processing time and also allows for better 3D processing. Further on, a pre-trained model based on the ImageNet dataset is used to increase the accuracy of the limited dataset. Then several methods, including CNNs, are applied to the pre-trained model. The optimization algorithms AdaDelta and Adam are used. The model achieves an accuracy of 66\%. However, when applying the pre-trained model, the accuracy becomes lower, and they conclude that they would need a larger dataset (>1200). Because of the lack of data, the authors therefore not conclude whether the use of transfer learning was successful, something this paper manages to do thanks to a larger dataset. 

\section{Conclusion}\label{sec:conclusion}

In this paper, we proposed an end-to-end machine learning model based on CNNs to translate images from the hand alphabet of SSL. We demostrate that the problem of having an small dataset of SSL data is solvable by using transfer-learning with a pre-trained model: our model is able to classify sign images with an accuracy of 85\%.

Our approach of using transfer-learning to boost the model accuracy might be replicable to many other sign languages. However, by using neural networks, the impact of human supervision is minimized and some patterns that are too complex for human's perception can be derived meaning from. However, the use of neural networks also adds an abstraction layer which humans have no control in understanding or changing. This means that the algorithm might do things it is not supposed to or derives too much meaning out of a binary situation. In this regard, the use of more traditional methods must not be neglected on the basis of new and more complex techniques. 

In the long run, this research could benefit deaf people who have access to technology and enhance good health, quality education, decent work, and reduced inequalities. Suggestions for future work include integrating dynamic signing data to interpret words and sentences, evaluating the method on another sign language's hand alphabet, and integrate dynamic interpretation in the web application for several letters or words to be interpreted in tandem. 

\section*{Acknowledgments}\label{sec:ak} 
This work is based on the Bachelor thesis of Peterson and Halvardsson \cite{Halvardsson2020}, which was conducted in collaboration with Prevas AB, a Swedish technical IT consulting firm focusing on several areas of industry such as energy, defence, and life science. It has also been partially supported by the Wallenberg Autonomous Systems and Software Program (WASP) funded by Knut and Alice Wallenberg Foundation.\looseness=-1

%% file: tables/pre-trained-models.tex
\begin{table}
\centering
\captionsetup{justification=centering}
\caption{The eleven pre-trained models available from Keras condidered in this paper for transfer learning. All models are trained on the ImageNet dataset. Each model is presented with its accuracy on the ImageNet dataset, and the number of parameters in the pre-trained model. The table is sorted on highest accuracy.}
~\\
\label{tab:pre-trained-models}
\begin{tabular}{lcc}             
\toprule
\textbf{Model} & \textbf{Accuracy $[$\%$]$} & \textbf{Parameters}\\
\midrule 
InceptionResNetV2 & 80.3 & 55,873,736\\   
Xception & 79.0 & 23,910,480\\   
InceptionV3 & 77.9 & 23,851,784\\  
ResNet50V2 & 76.0 & 25,613,800\\  
DenseNet121 & 75.0 & 8,062,504\\  
ResNet50 & 74.9 & 25,636,712\\  
NASNetMobile & 74.4 & 5,326,716\\  
MobileNetV2 & 71.3 & 3,538,984\\   
VGG16 & 71.3 & 138,357,544\\       
VGG19 & 71.3 & 143,667,240\\        
MobileNet & 70.4 & 4,253,864\\   
\bottomrule
\end{tabular}
\end{table}

%% file: tables/data-experiments.tex
\begin{table}
\centering
\captionsetup{justification=centering}
\caption{Information about the eight recordings included in the data generation. Each row represents a recording. The columns represent the person's age, gender, the total number of generated images as well as total time of video material for that recording. The eight recordings consisted of five different subjects, meaning recording one and six as well as two and eight were done by the same two persons respectively.}
~\\
\label{tab:data-experiments}
\begin{tabular}{ccccc} 
\toprule 
\textbf{Rec.} & \textbf{Age} & \textbf{Gender} & \textbf{\#Imgs} & \textbf{Rec. Time (s)}\\
\midrule
1 & 22 & Male & 5 191 & 519 \\
2 & 23 & Female & 5 200 & 520 \\ 
3 & 53 & Male & 5 200 & 520\\   
4 & 23 & Female & 4 401 & 440 \\ 
5 & 53 & Female & 5 200 & 520\\ 
6 & 22 & Male & 5 077 & 508\\  
7 & 24 & Male &  5 200 & 520\\ 
8 & 23 & Female & 5 134 & 513\\
\bottomrule
\end{tabular}
\end{table}

%% file: tables/accuracy-results.tex
\begin{table*}
\centering
\captionsetup{justification=centering}
\caption{Result, measured in accuracy, of the testing of pre-trained models and optimization algorithms. The columns correspond of the three pre-trained models tested, and the rows correspond to the two optimization algorithms. The results were obtained by basic layers added to the pre-trained models, and the new Swedish Sign Language data set.}
~\\
\label{tab:accuracy-results}
\begin{tabular}{lccc} 
\toprule
\multicolumn{1}{c}{\begin{tabular}[c]{@{}c@{}} \\ \end{tabular}} & \multicolumn{3}{c}{ \textbf{Pre-trained models}}  \\ 
\cline{2-4}
\textbf{Optimization algorithms}                                 & InceptionResNetV2 & Xception & InceptionV3        \\ 
\midrule
Adaptive Moment Estimation (Adam)                                & 88.51\%           & 94.77\%  & 91.77\%            \\ 
\midrule
Mini-batch gradient descent                                      & 87.98\%           & 91.63\%  & 94.78\%            \\
\bottomrule
\end{tabular}
\end{table*}